\documentclass[letterpaper, 10 pt, conference]{ieeeconf}  

\IEEEoverridecommandlockouts                              

\usepackage{cite}
\usepackage{amsmath,amssymb,amsfonts}
\usepackage{algorithmic}
\usepackage{graphicx}
\usepackage{textcomp}
\usepackage{xcolor}
\usepackage{enumerate}
\usepackage[linesnumbered,ruled,lined]{algorithm2e}
\usepackage{booktabs}
\usepackage{threeparttable}
\usepackage{multirow}
\usepackage{stfloats}
\usepackage{subfigure}
\usepackage{graphicx}
\usepackage{rotating}

\title{\LARGE \bf
Accurate and Robust Scale Recovery for Monocular Visual Odometry Based on Plane Geometry 
}

\author{Rui Tian$^{1}$, Yunzhou Zhang$^{1*}$, Delong Zhu$^{2*}$, Shiwen Liang$^{1}$, Sonya Coleman$^{3}$, Dermot Kerr$^{3}$
	\thanks{$^*$The corresponding author of this paper. }
	\thanks{$^{1}$Rui Tian, Yunzhou Zhang and Shiwen Liang are with College of Information Science and Engineering, Northeastern University, Shenyang 110819, China (Email: {\tt\small zhangyunzhou@mail.neu.edu.cn}).}%
	\thanks{$^{2}$Delong Zhu is with the Department of Electronic Engineering, The Chinese University of Hong Kong, Shatin, N.T., Hong Kong SAR, China (Email: {\tt\small zhudelong@link.cuhk.edu.hk}).}
	\thanks{$^{3}$Sonya Coleman and Dermot Kerr are with School of Computing and Intelligent Systems,Ulster University, N. Ireland, UK.}
	\thanks{This work was supported by National Natural Science Foundation of China (No. 61973066, 61471110) , Fundation of Key Laboratory of  Aerospace System Simulation(6142002200301), Fundation of Key Laboratory of Equipment Reliability(61420030302), Fundamental Research Funds for the Central Universities(N182608004, N2004022) and the Distinguished Creative Talent Program of Liaoning Colleges and Universities (LR2019027).}
}

\begin{document}

\maketitle
\thispagestyle{empty}
\pagestyle{empty}

\begin{abstract}
	Scale ambiguity is a fundamental problem in monocular visual odometry. Typical solutions include loop closure detection and environment information mining. For applications like self-driving cars, loop closure is not always available, hence mining prior knowledge from the environment becomes a more promising approach. In this paper, with the assumption of a constant height of the camera above the ground, we develop a light-weight scale recovery framework leveraging an accurate and robust estimation of the ground plane. The framework includes a ground point extraction algorithm for selecting high-quality points on the ground plane, and a ground point aggregation algorithm for joining the extracted ground points in a local sliding window. Based on the aggregated data, the scale is finally recovered by solving a least-squares problem using a RANSAC-based optimizer. Sufficient data and robust optimizer enable a highly accurate scale recovery. Experiments on the KITTI dataset show that the proposed framework can achieve state-of-the-art accuracy in terms of translation errors, while maintaining competitive performance on the rotation error. Due to the light-weight design, our framework also demonstrates a high frequency of 20 Hz on the dataset. 
\end{abstract}

\section{INTRODUCTION}

Monocular Visual Odometry (MVO) is a popular method for camera pose estimation, but due to the scale ambiguity \cite{song2015high,zhou2016reliable, wu2020eao,2009Absolute,2010scale}, the MVO system cannot provide real odometry data. Therefore, an accurate and robust scale recovery algorithm is of great significance in the application of MVO \cite{hawkeye-zhu}.
The key of scale recovery is to integrate absolute reference information, such as the gravity orientation from IMU \cite{qin2018vins} or the depth measured by Lidar \cite{zhang2014real, li2017hybrid}. The baseline of stereo cameras can also serve as such a reference \cite{mur2017orb}. However, these sensors are not always available in real-world applications. Moreover, a complicated sensor calibration and fusion process is needed to align the MVO system with other sensors. 

Another frequently used method for removing scale ambiguity is to take as reference the height of a mounted camera above the ground plane, which remains a stable signal during the navigation of vehicles. The idea is to estimate a ratio between the real camera height and the relative one calculated from image features. The ratio then can be used to recover the real scale. The advantages of this method are significant since it does not depend on other sensors and is with high feasibility. The method is also regarded as one of the most promising solutions in this research area.

Prior work like \cite{zhou2019ground,zhou2016reliable,song2015high,1211380} typically leverages the results of feature matching to calculate the homography matrix and then decompose the matrix to estimate the parameters of the ground plane, based on which the relative camera height can be obtained. The major deficiency of this method is that the decomposition is very sensitive to noises and multiple solutions exist, which requires additional operations to eliminate the ambiguity. Some other work like \cite{grater2015robust,1211380} chooses to directly fit the ground plane using feature points that lie on the ground, e.g., the center regions of a road. By removing low-quality image features outside the target region, the robustness is improved. However, the target region may be occluded sometimes, which interferes with the detection of image features, thus degrading the accuracy of scale recovery.

In recent work \cite{yin2017scale,andraghetti2019enhancing,xue2020toward,wagstaff2020self, wang2020tartanair}, deep learning based MVO algorithms are proposed, in which the camera pose with a real scale is directly predicted by the neural network in an end-to-end manner. Such methods have received much attention in recent years, but their generalization ability across different scenarios is very limited \cite{wang2020tartanair}. Some other deep learning based methods take scale recovery as an independent problem. For instance, DNet \cite{xue2020toward} is proposed to perform ground segmentation and depth regression simultaneously. Based on the predicted depth points within the ground region, a dense geometrical constraint is then formulated to help recover the scale. In \cite{wagstaff2020self}, a scale-recovery loss is developed based on the idea of enforcing the consistency between the known camera height and the predicted one. Constrained by this loss, the neural network can predict more accurate ego poses. Nonetheless, these methods usually require a large-scale training process, and the computational cost is prohibitively expensive.

In this paper, we propose a light-weight method for accurate and robust scale recovery based on plane geometry. The method includes an efficient Ground Point Extraction (GPE) algorithm based on Delaunay triangulation \cite{shewchuk1996triangle} and a Ground Points Aggregation (GPA) algorithm for aggregating ground points from consecutive frames. Based on these two algorithms, a large number of high-quality ground points are selected. For scale recovery, we first formulate a least-square problem to fit the ground plane and then estimate the relative camera height to calculate the real scale. By leveraging the high-quality points and a RANSAC-based optimizer, the scale can be estimated accurately and robustly. Benefiting from the light-weight design of the algorithms, our method can achieve a 20Hz running frequency on the benchmark dataset. 

The main contributions of this work are as follows:
\begin{itemize}
	\item We propose a GPE algorithm based on Delaunay triangulation, which can accurately extract ground points.  
	\item We propose a GPA algorithm that can effectively aggregate local ground points and perform robust optimization of the ground plane parameters.  
	\item Based on the proposed algorithms, we implement a real-time MVO system with accurate and robust scale recovery, aiming to reduce scale drift and provide accurate odometry in long-distance navigations without loop closure. 
\end{itemize}

\begin{figure*}[ht]
	\centering
	\includegraphics[height=6.3cm]{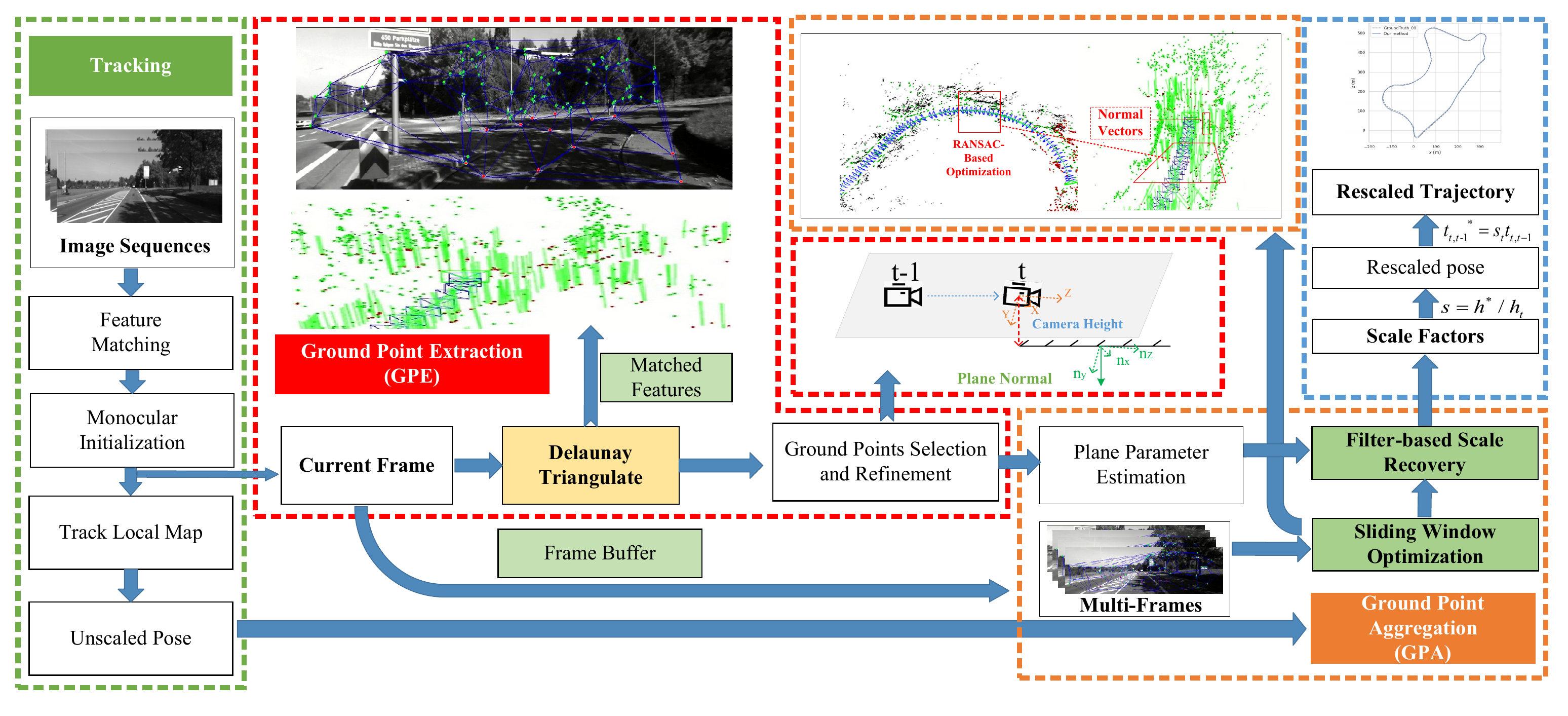}
	\caption{Overview of the proposed system. There are two parallel threads in the system: 1) The MVO thread takes image frames as input and estimates the current camera pose, 2) The GPE-GPA thread fetches image features from the MVO thread and selects high-quality points for ground plane estimation.}
	\label{overview}
	\vspace{-2mm}
\end{figure*}

\section{System Overview}
The notations used in this paper are as follows:
\begin{itemize}
	\item $T_t\in{R^{4\times4}}$ - The camera pose of image frame $I_t$ in the global frame, which is composed of a camera orientation $R_t\in{R^{3\times3}}$ and a translation $\boldsymbol{t}_t\in{R^{3\times1}}$.
	\item $T_{t,t-1}$ - The relative pose of frame $I_{t-1}$ w.r.t. frame $I_t$.
	\item $K$ - The intrinsic matrix of a pinhole camera model.
	\item $\mathbf{x}_i, \mathbf{u}_i$ - The 3D map point in camera frame and its corresponded 2D point on image plane after projection.
	\item $\mathbf{n}_t, h_t$ - The plane parameters, i.e.,  $\mathbf{n}_t \cdot \mathbf{x}_i-h_t=\mathbf{0}$, where $\mathbf{n}_t$ is the normal vector and $h_t$ is the distance to the plane.
	\item $h, {h}^{\dagger}, h^*$ -- The calculated camera height form image features, the estimated camera height after scale recovery, and the real camera height.
\end{itemize}

\subsection{Problem Definition}
Given consecutive image frames from a calibrated monocular camera, our goal is to estimate the absolute scale of camera poses and then recover the real camera trajectory by making use of the prior known camera height $h^*$. Under scale ambiguity, the camera height $h$ calculated from image features maintains a ratio with the real one, i.e., $s=h^*/h$. Therefore, scale recovery is essentially to compute $s$, and the key lies in the accurate estimation of the ground plane.

\subsection{System Architecture}
The proposed system in this work is shown in Fig. \ref{overview}. There are two parallel threads in the system: The first one is the MVO thread, which takes consecutive images as input and estimates the current camera pose, e.g., the ORB-SLAM2 framework. The second thread is used to run the GPE and GPA algorithms for scale recovery. The proposed system is based on such an assumption that the ground is locally flat and can be approximated by a plane with a surface normal. The workflow of the second thread is as follows.

As shown in the red block in Fig.\ref{overview}, for each image frame from the MVO thread, the Delaunay triangulation is first applied to segment the matched feature points into a set of triangles. Each triangle is then back-projected into the camera frame, and the associated plane parameters are also estimated. After that, several geometrical constraints are leveraged to select and then refine ground points. 

Note that selected ground points are not enough for an accurate estimation of the plane parameters. We thus propose the GPA algorithm to aggregate ground points from multiple frames using a sliding windows method, as shown in the orange block of Fig.\ref{overview}. Based on the aggregated local points, a robust parameter estimation procedure is then performed to fit the ground plane. Accordingly, the relative camera height of each frame can be estimated, and the absolute camera trajectory is recovered, shown in the blue block of Fig.\ref{overview}.

\section{Ground Plane Estimation}

\subsection{Ground Point Extraction}
For a given set of matched feature points $\mathbf{u}^t_i$, $i\in\{1,2,\dots,N\}$, in the current image frame $I_t$, the Delaunay triangulation uses each of the feature points as a triangle vertex. We back-project the triangles from the image plane into the current camera frame and denote them by $\Delta_i^t, i\in\{1,2,\dots M\}$ associated with a set of vertices $\mathbf{x}^t_{ij},j\in\{1,2,3\}$. The normal vector $\mathbf{n}_i^t = ({n}_{i,x}^t, {n}_{i,y}^t, {n}_{i,z}^t)$ of each triangle can be obtained by the cross product,
\begin{equation}
	\mathbf{n}_i^t=\frac{(\mathbf{x}^t_{i1}-\mathbf{x}^t_{i2})\times(\mathbf{x}^t_{i1}-\mathbf{x}^t_{i3})}{||(\mathbf{x}^t_{i1}-\mathbf{x}^t_{i2})\times(\mathbf{x}^t_{i1}-\mathbf{x}^t_{i3})||_2},
	\label{est-n}
\end{equation}
where $\mathbf{n}_i^t$ has an unit length. For each vertex of the triangle, the following geometrical constraint then holds:
\begin{equation}
	\mathbf{n}^t_i \cdot \mathbf{x}^t_{ij}-h^t_i=0.
	\label{est-h}
\end{equation}

Therefore, $h^t_i$ can also be estimated. Here, we also add two additional constraints, i.e., ${n}_{i,y}^t>0$ and $h^t_i>0$, based on the fact that the camera is mounted on the top of the vehicle and is above the ground plane, as shown in Fig.\ref{overview}.

Note that the triangles are scattered in the whole image plane, hence we need to identify the ones located on the ground plane, named \textit{ground triangles}, for estimating the plane parameters. Based on the fact that the normal of a ground triangles is orthogonal to camera translation $\boldsymbol{t}_{t,t-1}$, and that the pitch angle of the camera is zero, the ground triangles can be identified by testing with the following constraints,
\begin{equation}
	\begin{aligned}
		\arccos(\mathbf{n}^t_i, \boldsymbol{t}_{t,t-1}) &= 0, \\
		|\arctan(-\frac{R_{32}}{R_{33}})| = 0,\, &R_{33}\neq0. \\
	\end{aligned}
\end{equation}

In practice, the equality condition cannot be strictly satisfied. We thus set a tolerance value of $5^\circ$ in the actural implementation.

For ground triangles that satisfy the above constraints, their vertices are categorized into a new point set $\tilde{\mathbf{x}}^t_{ij}, i\in\{1,2,\dots K\}, j\in\{1,2,3\}, K\small{<}M$. Since the same vertex point may be shared by multiple triangles, we also need to remove the repetitive ones from the point set. This will ensure the same contribution of each point to the ground plane estimation. 

The ground points are now initially segmented out, denoted by $\tilde{\mathbf{x}}^t_{k} \in \mathcal{G}$, but there may still exist some outliers introduced by moving objects and some remote points.
To further improve the quality of $\mathcal{G}$, a RANSAC-based method is leveraged to optimize $\tilde{\mathbf{x}}^t_{k}$, which minimizes a plane-distance error as follows,
\begin{equation}
	\min_{\tilde{\mathbf{x}}^t_{g} \in \mathcal{G}} \;\sum_{g=1}^{|\mathcal{G}|}||\mathbf{n}^t\cdot\tilde{\mathbf{x}}^t_{g}-h^t||_2.
	\label{ransac}
\end{equation}

In the implementation, we randomly sample three points to estimate a new plane with eq. \eqref{est-n}-\eqref{est-h}, and then we calculate the total distance of the remaining points to the estimated plane. Such a process repeats $Z$ times, and the plane that induces the minimum total distance error is reserved. The points with a distance larger than $0.01$m to the reserved plane are then removed.
This is actually a more strict criterion for ground point selection. After this process, only high-quality ground points are reserved. In Alg. \ref{ag1}, we present a complete procedure of the GPE algorithm, which gives more details about the proposed implementation.

\begin{algorithm}[t]
	\caption{ Ground Point Extraction (GPE) }
	\label{ag1}
	\KwIn{  $\mathbf{u}^t_i$ , $R_{t,t-1}$, $\boldsymbol{t}_{t,t-1}$}
	\KwOut{$\{\tilde{\mathbf{x}}^t_{g}\}$}
	
	
	$\{\Delta_i^t\}$$\gets$ \textsc{DelaunayTriangulation}($\mathbf{u}^t_i$) 
	
	
	
	
	triangles points set $\{\tilde{\mathbf{x}}^t_{ij}\}\gets\emptyset$
	
	segmented points set $\{\tilde{\mathbf{x}}^t_{k}\}\gets\emptyset$
	
	ground points set $\mathcal{G}_{best}\gets\emptyset$
	
	temp ground points set $\mathcal{G}_k\gets\emptyset$

	\For{each  $\mathbf{u}_{ij}^t\in\{\Delta_i^t\}, j=\{1,2,3\}$ }{

		back-project $\mathbf{x}^t_{ij}$=$K^{-1}\mathbf{u}_{ij}^t$
		
		calculate $\mathbf{n}_i^t$ by Eq.(1)
		
		calculate $h_i^t$ by Eq.(2)
		
		\If{$	|\arctan(-\frac{R_{32}}{R_{33}})|<\theta_{1}$ \& $h^t_i>0$}
		{
			\If{$\arccos(\mathbf{n}^t_i, \boldsymbol{t}_{t,t-1})<\theta_{2}$}
			{

				$\{\tilde{\mathbf{x}}^t_{ij}\}\cup\mathbf{x}^t_{ij}$	
			}
		}	
	}

	delete repeate vertices and get $ \{\tilde{\mathbf{x}}^t_{k}\}\subset\{\tilde{\mathbf{x}}^t_{ij}\}$
	
	{/* \emph{ensure enough points} */} \\
	\While{size ($ \{\tilde{\mathbf{x}}^t_{k}\}$) $>$ 5 }{ 
		
		\For{iterations $<Z$}{
			randomly select  $\mathbf{x}^t_{1},\mathbf{x}^t_{2},\mathbf{x}^t_{3}\in \{\tilde{\mathbf{x}}^t_{k}\}$
			
			calculate $\mathbf{n}^t$ by Eq.(1)
			
			\For{$\tilde{\mathbf{x}}^t_{g}$ in $\{\tilde{\mathbf{x}}^t_{k}\}$}
			{
				$D\gets|\mathbf{n}^t\cdot\tilde{\mathbf{x}}^t_{g}-h^t|$ by Eq.(4) 
				
				\If{$D<d_{thresh}$}{
					$\mathcal{G}_k\cup \{\tilde{\mathbf{x}}^t_{g}\}$
				}
			}
			\If{size $(\mathcal{G}_k)$ $>$ size $(\mathcal{G}_{best})$ }
			{
				$\mathcal{G}_{best}\gets\mathcal{G}_k$ \tcp{best model select}
			}	
			
		}
		$\{\tilde{\mathbf{x}}^t_{g}\}\gets\mathcal{G}_{best}$
		
		\Return {$\{\tilde{\mathbf{x}}^t_{g}\}$}
	}
\end{algorithm}

\subsection{Ground Point Aggregation}

\begin{figure}[t]
	\centering
	\includegraphics[height=5.5cm]{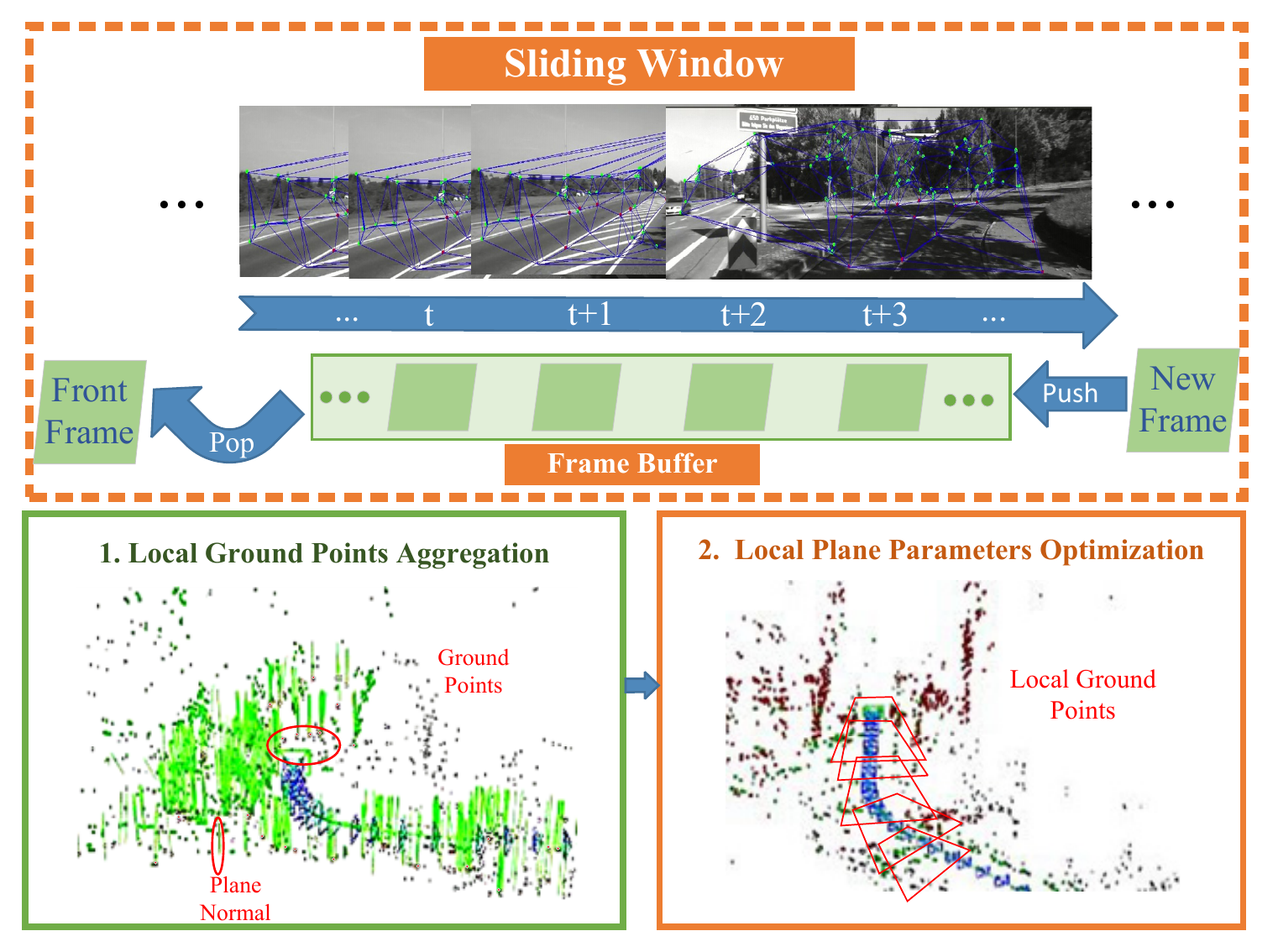}
	\caption{Illustration of the GPA algorithm. In the bottom-left figure, the red dots indicate ground points, and the green line segment is the normal of each triangle. In the bottom-right figure, the red quadrilateral is the estimated ground plane based on the aggregated ground points in the current window.}
	\label{lpg}
\end{figure}

Due to the strict criteria by GPE, the inliers are not enough for accurate estimation of the ground plane. Therefore, we propose the GPA algorithm to aggregate ground points from consecutive image frames. As shown in Fig. \ref{lpg}, we leverage the sliding window method to select image frames, and a frame buffer is maintained to store the camera poses and ground points in the current window. At each time step, with the arrival of a new image frame, we update the buffer and then estimate the ground plane by solving a least-squares problem.

From the MVO thread, we can get the pose $T_t$ and the inliers $\mathbf{x}^t_g$ of each frame from the buffer. We then can transform each of the inliers, denoted by $\mathbf{x}^t_{gi}$, into the global frame,
\begin{equation}
	\mathbf{p}_i=T_t \,[\mathbf{x}^t_{gi}, 1]^T.
\end{equation}

Suppose there are $N$ local ground points in the buffer, the least-squares problem that minimizes the plane-distance error of these points is formulated as follows:
\begin{equation}
	\begin{aligned}
		\min_{\boldsymbol{\mu}}\sum_{i=1}^{N}\|\boldsymbol{\mu}^T \mathbf{p}_i\|_2&=\min_{\boldsymbol{\mu}}\boldsymbol{\mu}{P_t}{P_t}^T\boldsymbol{\mu}^T, \\
		\boldsymbol{\mu} = &[\mathbf{n}_t, -h_t]^T,
	\end{aligned}
	\label{mat}
\end{equation}
where ${P_t}=[\mathbf{p}_1, \mathbf{p}_2, \cdots, \mathbf{p}_N ]\in R^{4\times N}$. Equation (\ref{mat}) can be rewritten as follows:
\begin{equation}
	\begin{aligned}
		&\min_{\boldsymbol{\mu}} \boldsymbol{\mu} Q_t\boldsymbol{\mu}^T,  \\ 
		Q_t&={P}_t{P}_t^T\in R^{4\times4},	
	\end{aligned}
	\label{opt}
\end{equation}
which can then be efficiently solved by the SVD method.

To further improve the estimation accuracy of $\boldsymbol{\mu}$, we also introduce a weighting matrix $\Sigma={\sigma^{-2}_z}I$, where $\sigma_z$ is the normalized distance of the point depth to their mean value. As a result, the matrix $Q$ in Eq. \eqref{opt} becomes,
\begin{equation}
	Q_t={P}_t \Sigma {P}_t^T\in R^{4\times4}.
\end{equation}

Another important refinement on the estimated plane parameter is to conduct a RANSAC-based optimization, which shares the same idea as Eq. \eqref{ransac}. In each iteration of the optimization, we first estimate $\boldsymbol{\mu}$, and then calculate the distance between $\mathbf{p}_i$ and the estimated plane. Points with a distance larger than $0.01$m are removed, and the remaining is then leveraged to estimate a new $\boldsymbol{\mu}$. Such a process continues until convergence. We denote the final plane normal by $\mathbf{n}_t^*$ and the reserved ground points by $\mathbf{p}_k^*, k\in\{1, 2, \cdots, K\}$. The relative camera height then can be calculated by projecting the camera center to the ground plane:
\begin{equation}
	h^t_j=\mathbf{n}_t^* \cdot (\mathbf{p}_c-\mathbf{p}_k^*),
	\label{multi-h}
\end{equation}
where $\mathbf{p}_c$ is the camera center of frame $I_t$ in the global frame. It is worth noting that there are $K$ estimated camera heights, which will be further processed to recover a smooth scale. Details of the GPA algorithm are presented in Alg. \ref{alg2}.

\begin{algorithm}[t]
	\label{alg2}
	\caption{Ground Points Aggregation (GPA)}
	
	\KwIn{ $I_t$, $\{\mathbf{x}^t_{g}\}$, $T_t$}
	
	\KwOut{$\{h^t\}$}
	
	buffer $\{queue\}\gets \emptyset$
	
	inliers in global frame $\{\mathbf{p}_t\} \gets \emptyset$
	
	inliers in current frame $\{\mathbf{x}_g^t\} \gets \emptyset$
	
	reserved ground points $\{\mathbf{p}_k^*\}\gets \emptyset$
	
	camera heights $\{h_t\}\gets \emptyset$

	\While{$I_t$ is not empty}{
		$\{queue\} \cup \{I_t\} $	
		
		\If{$size(\{queue\})>4$}{
			
			\textsc{pop$\_$front}($\{queue\}$) 
			
			\tcp{fixed number of frames}
			\For{$each$ $I_t$ in $\{queue\}$}{
				calculate $\mathbf{p}_i$ by Eq.(5)
				
				$\{\mathbf{p}_t\} \cup \{\mathbf{p}_i\} $
			}
			
			Matrix $P_t$
			
			calculate $\Sigma={\sigma^{-2}_z}I$ \tcp{Weighting Matrix}
			
			calculate $Q_t$ by Eq.(8)
			
			$\boldsymbol{\mu} \gets \textsc{SVDDecomposition}(Q)$
			
		}
		
		\For{iteration $<Z$}{
			Randomly select $\mathbf{p}_s \in \{\mathbf{p}_t\}$
			
			$D \gets \boldsymbol{\mu}^T \cdot \mathbf{p}_s $
			
			\If{$D > d_{thresh}$}{
				
				Remove $\mathbf{p}_s$ in $\{\mathbf{p}_t\}$

			}
			$\{\mathbf{p}_k^*\}\gets \{\mathbf{p}_t\}$
			
			calsulate $Q_s$ by Eq.(8)
			
			$Q \gets Q_s$
			
			$\boldsymbol{\mu_s} \gets \textsc{SVDDecomposition}(Q)$
			
			$\boldsymbol{\mu} \gets \boldsymbol{\mu_s}$ \tcp{update model}

		}
		
		\For{$\mathbf{p}^*$ in $\{\mathbf{p}_k^*\}$}{
			calculate $h_j^t$ by Eq.(9)
			
			$\{h^t\}\cup \{h_j^t\}$
		}
		\Return{$\{h^t\}$}
	}

\end{algorithm}

\subsection{Filter-Based Scale Recovery}
After we compute the relative camera height $h$ of each frame, the scale factor is then obtained by $s_t=h^*/h$, the motion scale of each frame is recovered by
\begin{equation}
	\boldsymbol{t}^{\dagger}_{t,t-1}=s_t \cdot \boldsymbol{t}_{t,t-1}.
\end{equation}

Corresponding to the multiple $h$ values in Eq. \eqref{multi-h}, there are also multiple estimated scales.

\begin{figure}[!htbp]
	\centering	
	\subfigure[The distribution of the scaled camera heights.]{
		\includegraphics[width=0.90\linewidth]{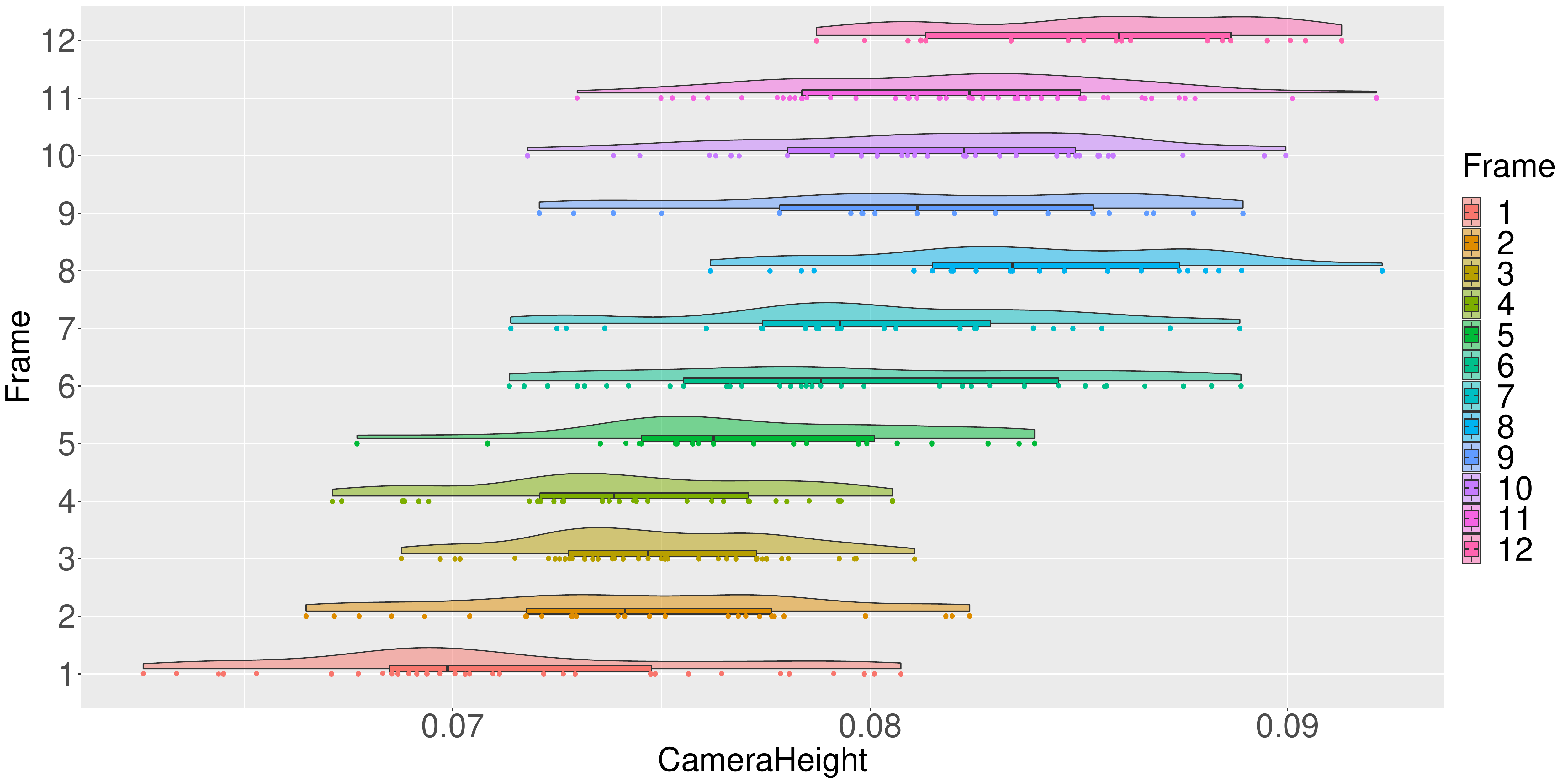}
		\label{gaussian}
	}
	\subfigure[The estimated camera height on sequence-02 and -05.]{
		\includegraphics[width=0.50\linewidth]{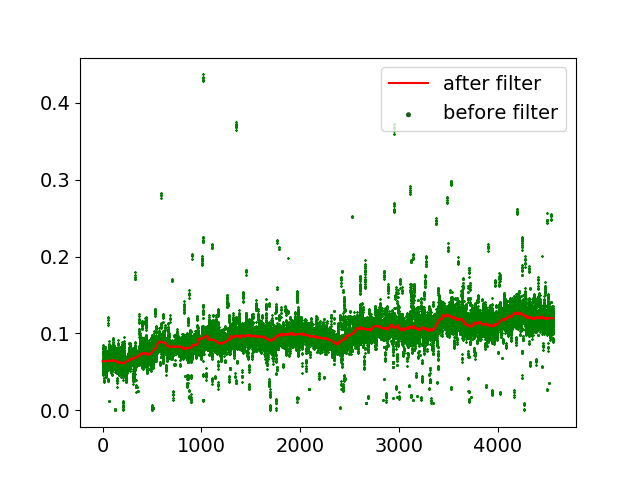}\hspace{-5mm}
		\includegraphics[width=0.50\linewidth]{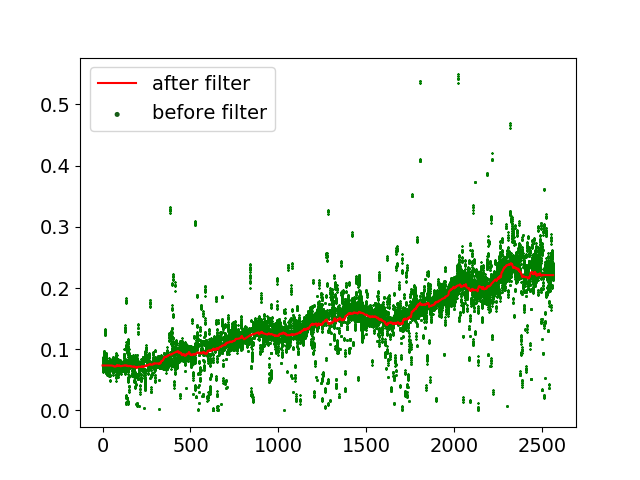}
		\label{filter-b}
	}	
	\centering
	\caption{Demonstration of the filter-based scale recovery. The green points are the scaled camera heights. The red curve is the smoothed one.}
	\label{filter}
\end{figure}

By plotting the scaled camera heights of each frame in the figure, shown in Fig. \ref{gaussian}, we find the data do not strictly follow a Gaussian distribution. Therefore, we choose the median point as the scale of the current frame. In the time domain, a moving average filter is applied, shown in Fig. \ref{filter-b}, which can give a more smooth result.

\section{Experiments}
We conduct experiments to evaluate the performance of our proposed method. The MVO system used in the experiments is implemented based on ORB-SLAM2, and the proposed scale recovery method is integrated as an independent thread. The system architecture is demonstrated in Fig. \ref{overview}. The KITTI dataset \cite{geiger2012we} is adopted as the benchmark dataset, in which sequence-01 is not used since it fails most feature-based VO systems. All the experiments are conducted using a laptop with Intel(R) Core(TM) i5-6300HQ CPU @ 2.30 GHz. 

\begin{figure*}[ht]
	\centering
	\includegraphics[height=8.2cm]{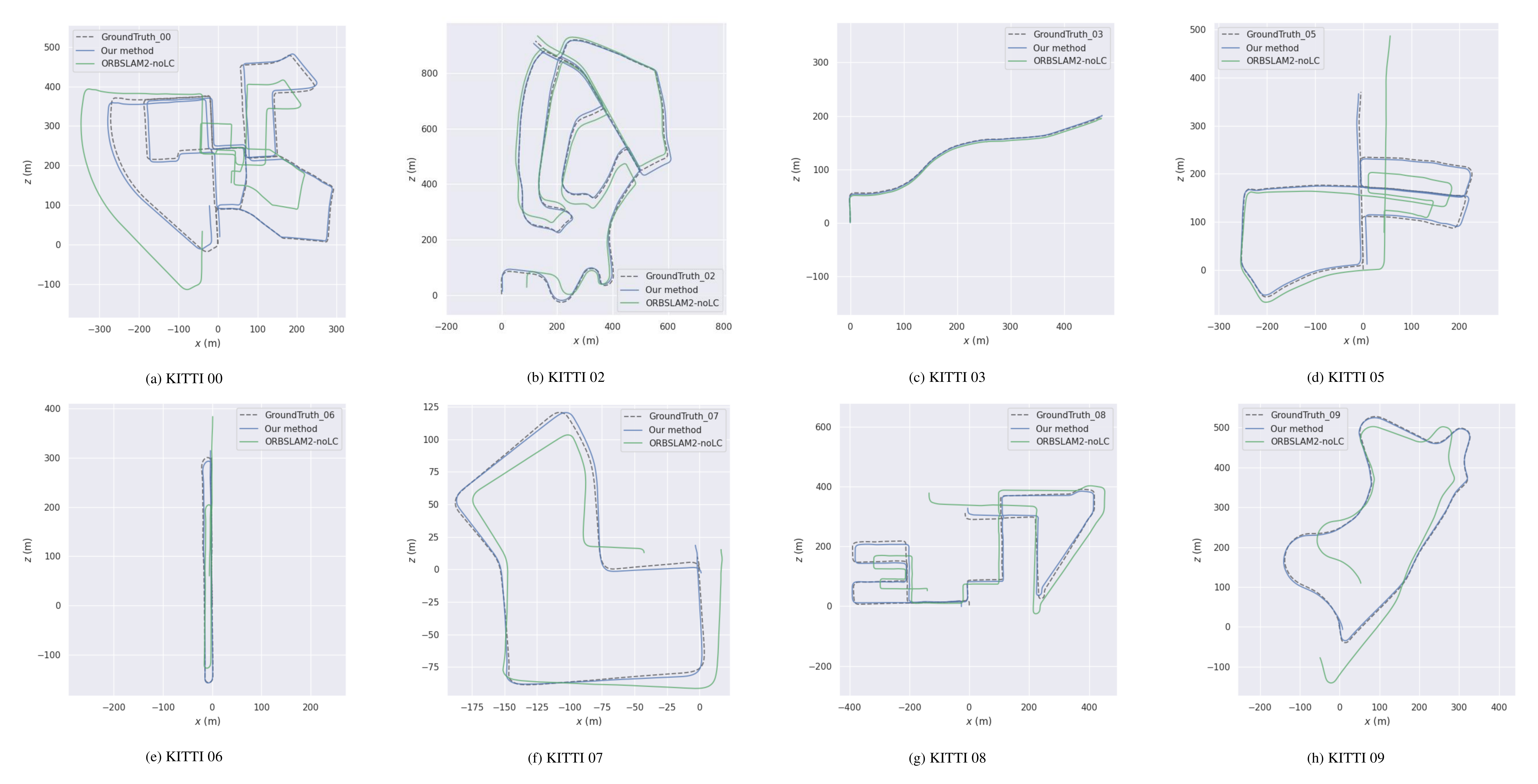}
	\caption{The re-scaled trajectories on KITTI dataset. The blue and green trajectories are generated by our system and ORB-SLAM2-noLC, respectively.}
	\label{fig3}
\end{figure*}

\subsection{Qualitative Evaluation}

The qualitative evaluation results of the proposed method are visualized in Fig. \ref{fig3}. The trajectories outputted by the system are recovered using the proposed method, which means similarity transformation is not necessary when compared with ground-truth trajectories\cite{grupp2017evo}. The baseline trajectories, indicated by green color in Fig. \ref{fig3}, are generated by ORB-SLAM2 with loop closure detection disabled, denoted by ORB-SLAM2-noLC. We can see that our re-scaled trajectories can eliminate scale drift and form correct loops, which demonstrates the effectiveness of the proposed method. 

The comparison of trajectory length between the ground truth and the proposed method is shown in Table \ref{tabel-1}, in which the Relative Length Error (RLE) is computed by $e=|l_{gt}-l_{ours}|/|l_{gt}|$, where $l$ is the length of a trajectory.
For sequence-02, -04, -06, -07, and -10, the RLE is less than 1$\%$. The high performance is due to the fact that most roads in these sequences are straight lines and contain rich features. Sequence-00 and -08 are more complicated cases, in which the scenario is composed of a lot of curves and turns. The path lengths are relatively longer than other sequences. The RLE is thus slightly higher, 2.17$\%$ and 2.72$\%$, respectively. Nevertheless, the results show that the proposed system can estimate an accurate scale of the entire trajectory.

\begin{table}[t]
	\renewcommand\tabcolsep{16.0pt} 
	\caption{Comparison of re-scaled trajectory \protect\\length with ground truth}
	\begin{center}
		\label{table_time}	
		\begin{tabular}{lllllll}
			\toprule
			Seq & GT (m)&Ours (m) & RLE (\%)\\
			\midrule
			00 & 3645.213 & 3724.419 & 2.173   \\
			02 &  5071.151 &5067.233 &0.757   \\
			03 &547.774 &560.888 &0.558 \\
			04 &391.861 &393.645 &0.455   \\
			05 &2175.159 &2205.576 &1.398   \\
			06 &1235.982 &1232.876 &0.251   \\
			07 &709.065 &694.697 &0.767   \\
			08 &3137.398 &3222.795 &2.722   \\
			09 &1738.601 &1705.051 &1.929   \\
			10 &911.399 &919.518 &0.890   \\
			\bottomrule  		
		\end{tabular}
	\end{center}
	\label{tabel-1}
\end{table}
\renewcommand{\arraystretch}{1} 
\begin{table*}[t]
	\centering
	\fontsize{6}{8.4}\selectfont
	\begin{threeparttable}
		\caption{comparison of average translation errors and rotation errors with the latest visual odometry methods on kitti dataset}
		\label{table2}
		\begin{tabular}{ccccccccccccccccccc}
			\toprule
			\multirow{3}{*}{Seq}& 	
			\multicolumn{2}{c}{ORB-SLAM2-noLC \cite{mur2017orb}}
			&\multicolumn{2}{c}{ VISO2-M\cite{2011StereoScan}}
			&\multicolumn{2}{c}{VISO2-Stereo\cite{2011StereoScan}}
			&\multicolumn{2}{c}{Song et.al\cite{song2015high}}
			&\multicolumn{2}{c}{Zhou et.al\cite{zhou2019ground}}
			&\multicolumn{2}{c}{Brandon et.al\cite{wagstaff2020self}}
			&\multicolumn{2}{c}{DNet\cite{xue2020toward}}
			
			&\multicolumn{2}{c}{Ours}\cr
			\cmidrule(lr){2-3} \cmidrule(lr){4-5}
			\cmidrule(lr){5-6} \cmidrule(lr){6-7}
			\cmidrule(lr){8-9} \cmidrule(lr){10-11}
			\cmidrule(lr){12-13} \cmidrule(lr){14-15}
			\cmidrule(lr){16-17}
			&Trans&Rot&Trans&Rot&Trans&Rot&Trans&Rot
			&Trans&Rot&Trans&Rot&Trans&Rot&Trans&Rot\cr
			&(\%)&(deg/m)&(\%)&(deg/m)&(\%)&(deg/m)&(\%)&(deg/m)
			&(\%)&(deg/m)&(\%)&(deg/m)&(\%)&(deg/m)&(\%)&(deg/m)\cr
			\midrule
			00&20.8&$-$&11.91&0.0209&2.32&0.0109&2.04&0.0048&2.17&0.0039&1.86& &1.94&$-$  &\textbf{1.41}&0.0054\cr
			
			01&$-$&$-$&$-$&$-$&$-$&$-$&$-$ &$-$ & $-$&$-$ &$-$ &$-$ &$-$&$-$&$-$&$-$\cr
			
			02&9.52&$-$&3.33&0.0114&\textbf{2.01}&0.0074&1.50&0.0035& $-$&$-$ &2.27&$-$ &3.07&$-$  &2.18&0.0046\cr
			
			03&$-$&$-$&10.66&0.0197&2.32&0.0107&3.37&0.0021&2.70&0.0044&$-$& $-$&$-$&$-$  &\textbf{1.79}&0.0041\cr
			
			04&$-$&$-$&7.40&0.0093&0.99&0.0081&2.19&0.0028&$-$ &$-$ &$-$&$-$ &$-$&$-$  &\textbf{1.91}&0.0021\cr
			
			05&18.63&$-$&12.67&0.0328&1.78&0.0098&1.43&0.0038&$-$ &$-$ &\textbf{1.50}&$-$ &3.32&$-$  &1.61&0.0064\cr
			
			06&18.98&$-$&4.74&0.0157&\textbf{1.17}&0.0072&2.09&0.0081& $-$& $-$&2.05&$-$ &2.74&$-$  &2.03&0.0044\cr
			
			07&13.82&$-$&$-$&$-$&$-$&$-$&$-$&$-$&$-$ &$-$ &1.78& $-$&2.74& $-$   &\textbf{1.77} &0.0230\cr
			
			08&22.06&$-$&13.94&0.0203&2.35&0.0104&2.37&0.0044& $-$& $-$&2.05&$-$ &2.72&$-$  &\textbf{1.51}&0.0076\cr
			
			09&12.74&$-$&4.04&0.0143&2.36&0.0094&1.76&0.0047&$-$ &$-$ &\textbf{1.50}&$-$ &3.70& $-$ &1.77&0.0118\cr
			
			10&4.86&$-$&25.20&0.0388&1.37&0.0086&2.12&0.0085&2.09&0.0054&3.70& &5.09&  &\textbf{1.25} &0.0031& \cr
			\midrule
			Avg&18.17& $-$&14.39&0.0245&2.32&0.0095&2.03&\textbf{0.0045}&2.32&0.045&2.03 & $-$&3.17&$-$&\textbf{1.72} & 0.0068&\cr
			
			\bottomrule
		\end{tabular}
	\end{threeparttable}
\end{table*}

\subsection{Quantitative Evaluation}

{The quantitative comparison} between our method and the baseline methods, including \cite{wagstaff2020self,xue2020toward,song2015high,zhou2019ground,2011StereoScan,mur2017orb}, is presented in Table \ref{table2}. The average translation error and rotation error are adopted as evaluation metrics.

We can see that ORB-SLAM2-noLC and VISO2-M have the worst performance due to the lack of loop closure detection. The scale drift of the two methods induces a large translation error, 18.17$\%$ and 14.39$\%$ respectively, while the VO systems with scale recovery all maintain a low translation error, $<4\%$. It can also be seen that a MVO system with scale recovery \cite{song2015high, zhou2019ground, xue2020toward, wagstaff2020self} can exhibit competitive performance with a stereo VO system like VISO2-M \cite{2011StereoScan}, which significantly demonstrates the importance of scale recovery for MVO. 

In terms of monocular systems, we can see our proposed method achieves the minimum translation error while maintaining a competitive performance on rotation error. The methods proposed by Song \textit{et al.} \cite{song2015high} and Zhou \textit{et al.} \cite{zhou2019ground} can not work with sequence-07, because they both rely on a fixed region to extract ground points, whereas occlusions by moving vehicles occur frequently in this sequence. In contrast with \cite{song2015high, zhou2019ground}, the proposed method works well with sequence-07 with a translation error of 1.77$\%$, benefiting from the GPA algorithm.

In \cite{xue2020toward}, a deep neural network, named DNet, is proposed for monocular depth prediction and scale recovery. Compared with this method, our method shows a better accuracy in all the sequences. In \cite{wagstaff2020self}, a real-time MVO system is implemented based on self-supervised learning techniques. This method can slightly outperform our proposed method in sequence-05 and -09, but has a much lower accuracy in sequence-00, -08, and -10. A similar phenomenon can be observed when comparing with \cite{song2015high}. This indicates a significant variance on the performance of \cite{wagstaff2020self}. Actually, this is the limitation of most deep learning based methods, which has been discussed in detail by \cite{wang2020tartanair}.    

The comparative experiments in Table \ref{table2} significantly verify the effectiveness of our method and demonstrates the advantages over the latest methods in the literature.
\subsection{Efficiency Evaluation}
\begin{figure}[ht]
	\centering
	\includegraphics[scale=0.45]{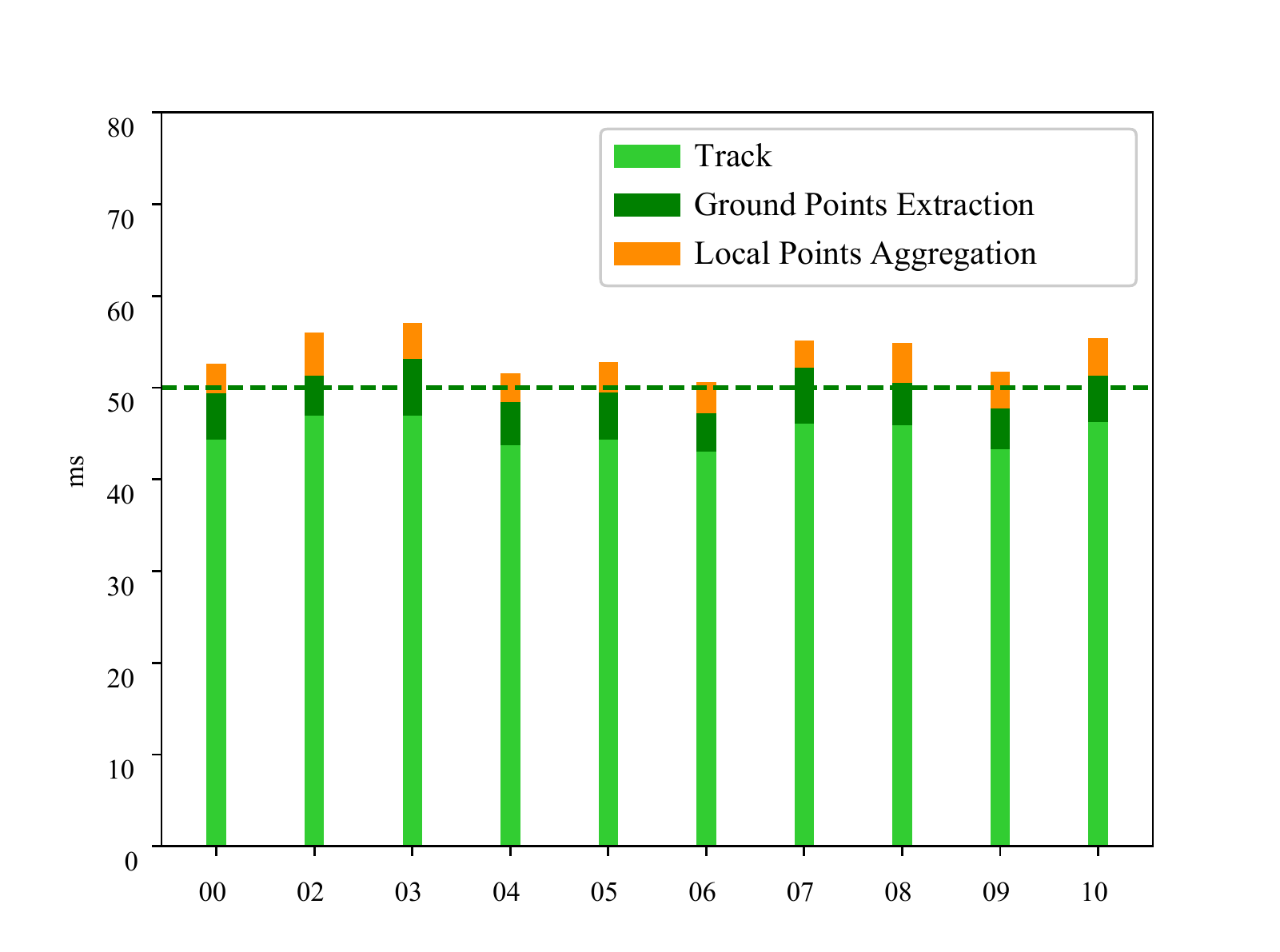}
	\caption{Run-time of our MVO system on KITTI dataset. The time costs of MVO thread, the GPE algorithm, and the GPA algorithm are reported.}
	\label{time}
\end{figure}

Another significant advantage of our method lies in its high efficiency. We evaluate the run-time of our system on all the KITTI sequences mentioned above, and the experiment repeats five times. The median run-time is reported in Fig. \ref{time}. In all the experiments, the MVO thread requires $50$-$55$ ms, while the GPE and GPA requires less than $10$ ms, which makes the system suitable for real-time applications.   
\section{CONCLUSIONS}

In this work, we present a light-weight MVO system with accurate and robust scale recovery, aiming to reduce scale drift and provide accurate odometry in long-distance navigations without loop closure. We solve the scale ambiguity for MVO by implementing our GPE-GPA algorithm for selecting high-quality points and optimizing them in a local sliding window. Sufficient data and robust optimizer provide accurate metric trajectory leveraging the ratio of the estimated camera height and the real one. Extensive experiments show that our proposed framework can achieve state-of-the-art accuracy and recover a metric trajectory without additional sensors. The system is designed to be a light-weight framework, which can achieve real-time performance with 20 Hz running frequency. Our proposed light-weight MVO system facilitates the localization and navigation of low-cost autonomous vehicles in long-distance scenarios.

Further study into integrating the uncertainty of plane estimation will be considered, which will further improve the accuracy of scale recovery. The light-weight neural network for ground segmentation will also be considered to help constrain the extraction of high-quality ground points.

\enlargethispage{-7.8cm}

\bibliographystyle{IEEEtran}
\bibliography{ref}



\end{document}